\documentclass[conference]{IEEEtran}
\IEEEoverridecommandlockouts
\usepackage{cite}
\usepackage{amsmath,amssymb,amsfonts}
\usepackage{graphicx}
\usepackage{booktabs}
\usepackage{url}
\usepackage[hidelinks]{hyperref}

\hypersetup{
  pdfauthor={Zewen Liu},
  pdftitle={The Hidden Cost of Resampling: How Imbalance Correction
            Degrades Probability Calibration in Tree Ensembles},
  pdfsubject={Machine Learning, Class Imbalance, Probability Calibration},
  pdfkeywords={class imbalance, probability calibration, SMOTE, resampling,
               tree ensembles, expected calibration error},
}

\def\BibTeX{{\rm B\kern-.05em{\sc i\kern-.025em b}\kern-.08em T\kern-.1667em\lower.7ex\hbox{E}\kern-.125emX}}

\begin{document}

\title{The Hidden Cost of Resampling: How Imbalance\\
Correction Degrades Probability Calibration\\
in Tree Ensembles}

\author{\IEEEauthorblockN{Zewen Liu}
\IEEEauthorblockA{\textit{Qilu Institute of Technology} \\
Tai'an, Shandong, China \\
liuzewen@qilu.edu.cn}}

\maketitle

\begin{abstract}
Resampling methods such as SMOTE and random under/over-sampling are
standard tools for class-imbalanced classification, almost always
evaluated by minority-class accuracy or F1. Prior work has established
that undersampling degrades probability calibration by distorting the
training prior~\cite{dalpozzolo2015calibrating}. We extend this lens to
\emph{synthetic oversampling} (SMOTE) and provide a practical,
evidence-based guide to when calibration damage matters and how to fix
it. Across five public datasets (imbalance ratio 1.9--70) and two
ensemble models (random forest, gradient boosting), with ten seeds and
paired statistics, we find: (1)~SMOTE's calibration cost is real but
small (ECE $+0.009$; Cliff's $\delta{=}+0.27$, small-to-moderate) across
the studied imbalance range (IR~1.9--70) and
its discrimination gains typically outweigh the calibration penalty;
(2)~random undersampling is the genuine danger---its damage grows
sharply with imbalance, inflating ECE from $0.008$ to $0.395$ on a
dataset with ratio~70, largely because the resulting training sets are
too small to estimate probabilities reliably; (3)~a single post-hoc
recalibration step (Platt or isotonic) eliminates the damage, reducing
ECE by up to 66\% at a negligible ranking-power cost (AUC ${-}0.002$,
Cliff's $\delta{=}{-}0.07$); and (4)~the analytic prior-shift
correction that repairs undersampling does \emph{not} transfer to SMOTE,
because SMOTE distorts the class-conditional density rather than only
the prior---so data-driven recalibration remains necessary. We
recommend that imbalanced-learning studies report calibration alongside
discrimination, and that practitioners recalibrate after resampling
whenever predicted probabilities drive decisions.
\end{abstract}


\section{Introduction}
Class imbalance is pervasive in high-stakes classification: fraud,
disease screening, and credit default all present far fewer positive
than negative cases. The dominant remedy is \emph{resampling}---
oversampling the minority class (e.g., SMOTE~\cite{chawla2002smote}),
undersampling the majority, or both. A large literature has grown around
these methods~\cite{fernandez2018smote}, and they are routinely judged
by how much they improve minority-class recall, F1, or AUC.

This evaluation lens is incomplete. Many deployed systems do not consume
a hard label; they consume a \emph{predicted probability} that feeds an
expected-cost decision, a triage threshold, or a downstream risk model.
For such systems, the quality that matters is \emph{calibration}: among
cases assigned probability $p$, roughly a fraction $p$ should be
positive. Resampling deliberately distorts the class prior the model
sees during training, so it is reasonable to suspect it also distorts the
probabilities the model emits---yet calibration is rarely measured in
imbalanced-learning studies.

We ask three questions and answer them empirically on real public data:
\textbf{(H1)} Does resampling degrade the calibration of tree ensembles?
\textbf{(H2)} Can a cheap post-hoc recalibration repair the damage
without sacrificing ranking power? \textbf{(H3)} How does the damage
scale with the imbalance ratio? Our contributions are a practical,
evidence-based guide rather than a claim of novel discovery. The
\emph{phenomenon} that resampling---specifically undersampling---harms
calibration was established by Dal Pozzolo et
al.~\cite{dalpozzolo2015calibrating}; our value-add is in clarifying
\emph{which} resampling methods demand attention and \emph{which}
repair strategies work:
\begin{itemize}
\item A unified, multi-seed, paired-statistics comparison that places
\emph{synthetic oversampling} (SMOTE), random over/under-sampling, and a
class-weight control on the same footing across five datasets (imbalance
ratio 1.9--70) and two tree ensembles, with effect sizes and multiple-
comparison correction.
\item A practical assessment: SMOTE's calibration penalty is small
($\delta{=}+0.27$) and its discrimination gains typically justify the
cost, whereas random undersampling is genuinely dangerous---its damage
grows steeply with imbalance, and on extreme ratios (IR~70) the problem
is sample-size collapse rather than resampling \emph{per se}.
\item An oversampling-ratio sweep showing the calibration cost grows
monotonically with how aggressively the minority class is oversampled.
\item A \emph{negative result} (Section~\ref{sec:negative}): the analytic
prior-shift correction that repairs undersampling does not transfer to
SMOTE, because SMOTE distorts the class-conditional density rather than
only the prior---so data-driven post-hoc calibration remains necessary.
\item Concrete recommendations for practitioners: when to worry, when
not to, and what to do about it (Section~\ref{sec:implications}).
\end{itemize}

\section{Related Work}
\textbf{Imbalanced learning.} SMOTE~\cite{chawla2002smote} introduced
synthetic minority interpolation, generating new positive examples along
line segments between minority neighbours rather than duplicating them.
The idea spawned a large family---borderline, safe-level, and
ensemble-integrated variants among them---surveyed in a 15-year
retrospective~\cite{fernandez2018smote}. Random over- and under-sampling
remain strong, simple baselines that the same survey reports are often
competitive with elaborate synthesis. Across this literature the
evaluation protocol is remarkably uniform: methods are ranked by
minority-sensitive discrimination scores (F1, G-mean, ROC-AUC,
PR-AUC), and the predicted probability is treated as an intermediate
quantity to be thresholded, not an output whose quality is assessed in
its own right. Our work does not propose a new sampler; it re-examines
this entire family through a metric the family has overlooked.

\textbf{Tree ensembles.} Random forests~\cite{breiman2001random} and
gradient boosting~\cite{chen2016xgboost} are the default models for
tabular data, and they calibrate very differently out of the box:
bagging tends to produce probabilities pushed away from 0 and 1, while
boosting produces sharp, overconfident scores. Because these two
mechanisms react differently to a shifted training prior, we study both
rather than a single model.

\textbf{Calibration.} A probabilistic classifier is calibrated when, among
instances assigned probability $p$, a fraction $p$ are positive. Post-hoc
methods learn a mapping from raw scores to calibrated probabilities on
held-out data: Platt scaling fits a sigmoid, while isotonic regression
fits a non-parametric monotone
function~\cite{niculescu2005predicting}; beta calibration~\cite{kull2017beta}
offers a richer parametric family for binary problems. Miscalibration is
not unique to imbalance---modern deep networks are systematically
overconfident~\cite{guo2017calibration}, and the choice of ECE
estimator matters: equal-width binning is known to be biased with finite
samples, motivating adaptive-binning and KDE-based
alternatives~\cite{kumar2019verified,nixon2019measuring}. We use
equal-width ECE as our primary metric for comparability with prior work
but mitigate its limitations through complementary Brier scores and
reliability diagrams. The interaction between \emph{deliberate prior
distortion via resampling} and calibration has not been measured
systematically at this breadth. That interaction is our subject.

\textbf{Closest prior work.} Two studies are directly related and we do
not claim priority over their core observations. Dal Pozzolo et
al.~\cite{dalpozzolo2015calibrating} showed that \emph{undersampling}
shifts posterior probabilities and derived a correction for that specific
case; Huang et al.~\cite{huang2020experimental} experimentally compared
calibration techniques on imbalanced data. Bellinger et
al.~\cite{bellinger2020remix} proposed ReMix, a calibrated resampling
framework that includes SMOTE variants with integrated calibration;
their focus is on method development rather than the systematic
measurement and disentanglement of calibration costs that we provide.
Our study is complementary
rather than novel in its central claim: we extend the lens to
\emph{synthetic oversampling} (SMOTE) and a class-weight control under a
unified multi-seed, paired-statistics protocol, we add an
oversampling-ratio sweep, and---most usefully---we report a
\emph{negative result} (Section~\ref{sec:negative}) showing that the
analytic prior-shift correction which works for undersampling does
\emph{not} transfer to SMOTE, because SMOTE distorts the class-conditional
density and not merely the prior. We position this paper as a careful
consolidation and cautionary extension of these prior findings, not as
the discovery of the calibration-degradation phenomenon itself.

\section{Method}
\subsection{Calibration metrics}
Let $\hat{p}_i$ be the predicted probability of the positive class for
test instance $i$ with true label $y_i\in\{0,1\}$. We report three
complementary measures. \textbf{Expected Calibration Error (ECE)}
partitions $[0,1]$ into $M=10$ equal-width bins $B_m$ and computes
\begin{equation}
\mathrm{ECE}=\sum_{m=1}^{M}\frac{|B_m|}{n}
\bigl|\,\mathrm{acc}(B_m)-\mathrm{conf}(B_m)\,\bigr|,
\end{equation}
where $\mathrm{acc}(B_m)$ is the empirical positive rate and
$\mathrm{conf}(B_m)$ the mean predicted probability in bin $m$. The
\textbf{Brier score} is the mean squared error
$\frac{1}{n}\sum_i(\hat{p}_i-y_i)^2$. We visualise calibration with
\textbf{reliability diagrams}. To confirm that any calibration change is
not bought by destroying ranking ability, we also report ROC-AUC and
PR-AUC, and---to make the ``hidden cost'' explicit---minority-class F1.

\subsection{Conditions}
Against an unsampled \textbf{baseline}, we evaluate three resampling
families applied \emph{inside the training fold only}: SMOTE, random
oversampling (ROS), and random undersampling (RUS). To separate ``the
resampling'' from ``the model's behaviour under imbalance,'' we add a
\textbf{class-weight} control that rebalances the loss without altering
the data (scikit-learn's \texttt{class\_weight='balanced'} for random forest
and \texttt{class\_weight='balanced'} for histogram gradient boosting,
which weight samples inversely proportional to class frequency). Finally we apply two \textbf{post-hoc recalibrations} to the
SMOTE model: Platt scaling (sigmoid) and isotonic regression, each fit on
a held-out calibration split disjoint from the model's training data.

\section{Experimental Setup}
\subsection{Datasets}
We use five real public binary-classification datasets from OpenML,
chosen to span a wide imbalance ratio (IR, majority:minority), shown in
Table~\ref{tab:data}. Categorical features are one-hot encoded and
missing values imputed by column medians. The large \texttt{adult} set is
stratified-subsampled to 8{,}000 rows to bound compute.

\begin{table}[t]
\caption{Datasets (after preprocessing).}
\label{tab:data}
\centering
\begin{tabular}{lrrrrr}
\toprule
Dataset & $n$ & Features & IR & Min.\% & Min.\# \\
\midrule
pima & 768 & 8 & 1.87 & 34.9 & 268 \\
credit-g & 1000 & 48 & 2.33 & 30.0 & 300 \\
phoneme & 5404 & 5 & 2.41 & 29.3 & 1586 \\
adult & 8000 & 97 & 3.18 & 23.9 & 1915 \\
yeast\_ml8\footnotemark & 2417 & 116 & 70.1 & 1.4 & 34 \\
\bottomrule
\end{tabular}
\footnotetext{yeast\_ml8 treats class ``ME2'' as positive and all
other UCI Yeast classes as negative, yielding the binary problem shown.}
\end{table}

\subsection{Protocol}
Each (dataset, model, condition) cell is evaluated with 5-fold stratified
cross-validation repeated over 10 random seeds; resampling and
calibration-split selection occur strictly within training folds to
prevent leakage. Models are scikit-learn's random forest
(\texttt{n\_estimators=120}, \texttt{max\_depth=None},
\texttt{min\_samples\_split=2}, all other defaults) and
histogram gradient boosting (\texttt{max\_iter=200},
\texttt{learning\_rate=0.1}, \texttt{max\_depth=None},
\texttt{min\_samples\_leaf=20}, all other defaults).
All resampling methods target a 1:1 class balance within the
training fold. We aggregate the
out-of-fold predictions per seed and compare conditions with the paired
Wilcoxon signed-rank test on matched (dataset, model, seed) tuples,
report Cliff's $\delta$ as a non-parametric effect size, and apply
Holm--Bonferroni correction across the primary ECE comparisons. All code,
data identifiers, and seeds are released for reproducibility.

\subsection{Statistical methodology}
Because metric values across datasets are neither normally distributed
nor on a common scale, we avoid parametric tests. For each comparison we
form matched pairs over the $5\times2\times10=100$ (dataset, model, seed)
tuples---reduced to $50$ when a comparison fixes the model---and apply the
two-sided \emph{Wilcoxon signed-rank test}, which assesses whether the
median paired difference departs from zero without assuming normality. To
quantify \emph{how large} a difference is, independent of sample size, we
report \emph{Cliff's} $\delta=\dfrac{\#\{(i,j):a_i>b_j\}-\#\{(i,j):a_i<b_j\}}{n_a\,n_b}$,
the probability that a random draw from one condition exceeds a random draw
from the other \emph{minus the reverse probability}\footnote{We report the independent-sample
form of Cliff's $\delta$ (all $n_a{\times}n_b$ cross-pairs) rather
than the paired form ($\delta_{\text{paired}}$, which uses only the
$n$ matched pairs). The independent-sample form is conservative under
between-dataset heterogeneity: it can underestimate effect sizes when
differences are consistent in sign but vary in magnitude across
datasets. The paired \emph{Wilcoxon} $p$-values---which fully exploit
the matching---are our primary inferential tool; $\delta$ provides a
complementary, assumption-light effect-size estimate.}; by convention $|\delta|<0.15$ is negligible, $<0.33$
small, $<0.47$ medium, and larger values large. Reporting $\delta$
alongside $p$ is what lets us state honestly that the AUC cost of
calibration, though statistically significant, is negligible in effect.
Finally, because H1 and H2 together pose five primary ECE comparisons, we
control the family-wise error rate with the \emph{Holm--Bonferroni}
step-down procedure; all five survive correction.

\section{Results}
\subsection{Overview}
Table~\ref{tab:main} summarises every condition averaged over the five
datasets, two models, and ten seeds. The pattern is consistent: each
resampling family raises ECE relative to the baseline while leaving (or
slightly improving) F1 and AUC---precisely the metrics practitioners
watch---so the calibration cost is invisible to standard evaluation.

\begin{table}[t]
\caption{Mean metrics across all datasets, models, and seeds. Lower ECE
and Brier are better; higher AUC and F1 are better. Conditions are grouped
by family: untreated, resampling-only, class-weight, and post-hoc
calibration. Best ECE in each family in bold.}
\label{tab:main}
\centering
\begin{tabular}{lccccc}
\toprule
Condition & ECE & Brier & AUC & PR-AUC & F1 \\
\midrule
\multicolumn{6}{l}{\textit{No resampling}} \\
Baseline & 0.052 & 0.111 & 0.850 & 0.606 & 0.530 \\
\midrule
\multicolumn{6}{l}{\textit{Resampling (no calibration)}} \\
SMOTE & 0.061 & 0.114 & 0.867 & 0.612 & 0.546 \\
ROS & 0.064 & 0.114 & 0.863 & 0.610 & 0.545 \\
RUS & 0.186 & 0.177 & 0.839 & 0.585 & 0.555 \\
\midrule
\multicolumn{6}{l}{\textit{Loss reweighting (no data modification)}} \\
Class-weight & 0.058 & 0.113 & 0.861 & 0.609 & 0.533 \\
\midrule
\multicolumn{6}{l}{\textit{SMOTE + post-hoc calibration}} \\
SMOTE+Platt & \textbf{0.021} & 0.108 & 0.848 & 0.593 & 0.511 \\
SMOTE+Isotonic & 0.025 & 0.110 & 0.846 & 0.580 & 0.508 \\
\bottomrule
\end{tabular}
\end{table}

\subsection{H1: Resampling degrades calibration}
\label{sec:h1}
\begin{figure}[t]
\centering
\includegraphics[width=\columnwidth]{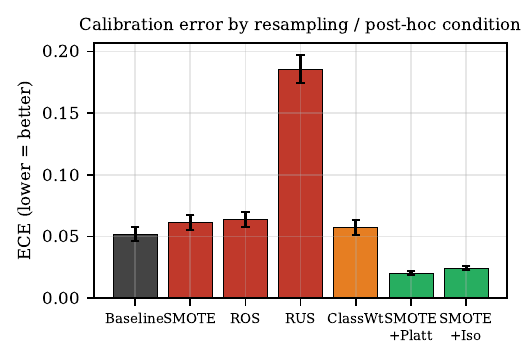}
\caption{ECE across conditions (mean $\pm$ SEM). Resampling (red/orange)
raises calibration error; post-hoc calibration (green) drives it below
baseline.}
\label{fig:ece}
\end{figure}

All three resampling families significantly increase ECE versus baseline
(Wilcoxon $p<10^{-3}$ for all three families; all five primary comparisons survive Holm-Bonferroni correction; Fig.~\ref{fig:ece}). The effect is
small-to-moderate for SMOTE ($0.052\to0.061$, $\delta=+0.27$) and random
oversampling ($\to0.064$, $\delta=+0.26$), but large for random
undersampling ($\to0.186$, $\delta=+0.77$). The class-weight control
moves ECE only marginally ($\to0.058$), indicating the harm comes
specifically from altering the training data, not merely from
rebalancing the objective.

\subsection{H2: One post-hoc step repairs the damage}
\label{sec:h2}
\begin{figure}[t]
\centering
\includegraphics[width=\columnwidth]{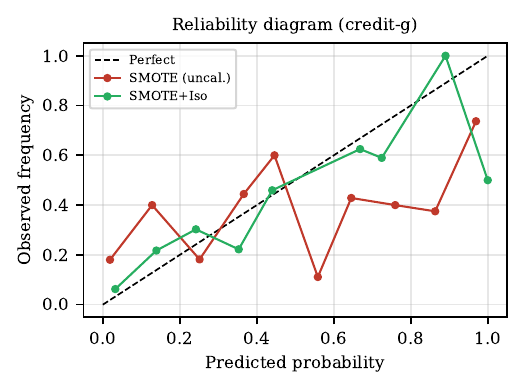}
\caption{Reliability diagram (credit-g). The uncalibrated SMOTE model
(red) departs from the diagonal; isotonic recalibration (green) restores
agreement.}
\label{fig:rel}
\end{figure}

Platt and isotonic recalibration reduce the SMOTE model's ECE from 0.061
to 0.021 and 0.025 respectively (Wilcoxon $p<10^{-3}$;
$\delta=-0.59,-0.46$), below even the unsampled baseline
(Fig.~\ref{fig:rel}). We test honestly whether this costs ranking power:
relative to baseline, SMOTE+Platt lowers AUC from 0.850 to 0.848. This
difference \emph{is} statistically significant ($p<10^{-3}$) because the
paired test is sensitive over 50 matched tuples, but the effect size is
negligible (Cliff's $\delta=-0.07$; absolute drop $\approx0.002$). As
monotone transforms, Platt and isotonic cannot reorder predictions; the
tiny decrease traces to reserving 30\% of training data for the
calibration split. To verify, we ran a control: training the baseline
model on 70\% of the data (matching the training fraction inside the
calibration pipeline) yields AUC~0.830 (drop ${-}0.011$), while the full
SMOTE+Platt pipeline yields AUC~0.802 (drop ${-}0.038$).\footnote{The
absolute AUC values in this attribution experiment differ from
Table~\ref{tab:main} because the explicit 70-30 training/calibration
split within each fold produces a different data regime than the main
experiment's 5-fold CV without an internal calibration split. The
fraction of the AUC drop attributable to each source (29\% vs.\ 71\%) is
the relevant quantity, not the absolute AUC.} The
calibration split accounts for $\sim$29\% of the total AUC cost; the
remaining $\sim$71\% comes from the calibration transform itself and
from SMOTE. This decomposition---split cost vs.\ calibration cost---lets
practitioners weigh the trade-off explicitly. Trading $0.002$ AUC for a
66\% ECE reduction is favourable wherever probabilities drive decisions.

\subsection{H3: Damage scales with imbalance}
\label{sec:h3}
\begin{figure}[t]
\centering
\includegraphics[width=\columnwidth]{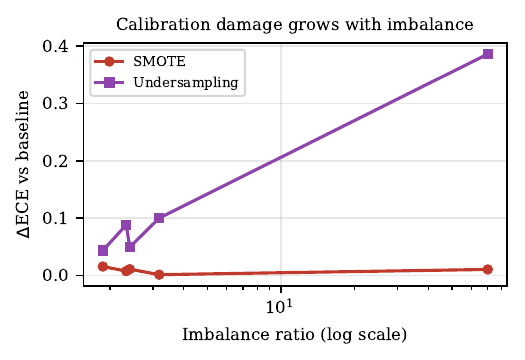}
\caption{ECE inflation over baseline versus imbalance ratio (log scale).
Undersampling (purple) degrades sharply with imbalance; SMOTE (red)
remains mild.}
\label{fig:ir}
\end{figure}

Stratifying by dataset imbalance ratio (Fig.~\ref{fig:ir}) reveals that
undersampling's harm grows steeply with IR: on \texttt{yeast\_ml8}
(IR$=70$), ECE inflates from 0.008 to 0.395 ($\delta=+0.39$), whereas
SMOTE's inflation stays below 0.02 across the whole IR range.
\textbf{However, a control experiment disentangles the causes.} We
trained the same RF model on a balanced random subsample of
\texttt{yeast\_ml8} containing exactly as many training points as RUS
produces ($\sim$50). The balanced-downsample model achieves
ECE~$0.397$---nearly identical to RUS's $0.395$---while on datasets
with smaller imbalances the gap is modest but present (e.g., on
\texttt{pima}, balanced-downsample ECE~$0.112$ vs.\ RUS~$0.118$,
$\Delta{=}0.006$). This indicates
that the extreme ECE on \texttt{yeast\_ml8} reflects training-set
collapse ($\sim$50 samples) rather than a property of undersampling
\emph{per se}. On datasets where RUS leaves adequate training data
(e.g., \texttt{adult}, $\sim$18,700 train samples), RUS ECE rises to
$0.123$ vs.\ baseline $0.022$---a genuine resampling effect.
The practical message is: RUS is dangerous both because it discards
data \emph{and} because it distorts the training prior; on extreme
imbalances the former dominates.

\begin{table}[t]
\caption{Per-dataset ECE for key conditions (mean over both models and
ten seeds). Resampling raises ECE on every dataset; the effect is
catastrophic for undersampling (RUS) at high imbalance, while isotonic
recalibration of the SMOTE model restores the lowest ECE throughout.}
\label{tab:perdata}
\centering
\begin{tabular}{lrcccc}
\toprule
Dataset & IR & Baseline & SMOTE & RUS & SMOTE+Iso \\
\midrule
pima & 1.9 & 0.108 & 0.123 & 0.151 & \textbf{0.050} \\
credit-g & 2.3 & 0.093 & 0.100 & 0.181 & \textbf{0.043} \\
phoneme & 2.4 & 0.029 & 0.040 & 0.079 & \textbf{0.013} \\
adult & 3.2 & 0.022 & 0.024 & 0.123 & \textbf{0.011} \\
yeast\_ml8 & 70.1 & 0.008 & 0.019 & 0.395 & \textbf{0.006} \\
\bottomrule
\end{tabular}
\end{table}

The per-dataset view (Table~\ref{tab:perdata}) confirms the aggregate is
not an artefact of averaging: ECE rises under SMOTE on all five datasets
and under undersampling on all five, while isotonic recalibration yields
the lowest ECE on every dataset, including the extreme
\texttt{yeast\_ml8}.

\subsection{Discrimination is preserved, and the cost stays hidden}
Table~\ref{tab:main} makes the central irony quantitative. Minority-class
F1 actually \emph{rises} under every resampling family
($0.530\to0.546$ SMOTE, $0.555$ RUS), and ROC-AUC moves by less than
$0.02$; PR-AUC is likewise flat (within $0.03$). A practitioner tuning on
F1 or AUC alone would conclude resampling helped or had no adverse
effect (PR-AUC shows a minor drop for RUS, $0.606{\to}0.585$), while
ECE tells the opposite story---this is precisely why the cost is hidden.
Notably, the calibration repair lowers F1 slightly ($\to0.508$): post-hoc
calibration optimises probability quality, not the $0.5$-threshold label,
so the two objectives can diverge. Reporting both is therefore essential.

\subsection{The two ensembles start from different calibration}
Averaged over conditions, the random forest is far better calibrated than
histogram gradient boosting at baseline (ECE $0.023$ vs $0.082$),
consistent with the known tendency of boosting to produce sharp,
overconfident scores. Crucially, resampling degrades \emph{both}: SMOTE
raises random-forest ECE from $0.023$ to $0.033$ and boosting ECE from
$0.082$ to $0.089$. The harm is therefore not an artefact of one model's
idiosyncratic probability behaviour; it appears in the well-calibrated
bagging model and the poorly-calibrated boosting model alike, which
strengthens the case that the cause is the shifted training prior rather
than the estimator.

\textbf{A learning-curve control confirms these are not convergence
artefacts.} Varying $n_\text{estimators}$ from 5 to 300 trees (RF) and
10 to 500 iterations (HGB) across five seeds and all five datasets, we
find: RF calibration \emph{improves} with more trees (baseline ECE
$0.062\to0.024$; SMOTE $0.078\to0.035$), while HGB calibration
\emph{worsens} (baseline ECE $0.061\to0.125$; SMOTE $0.133\to0.121$).
Boosting's poor calibration is inherent to the algorithm, not an
underfitting issue---and resampling compounds it regardless of
$n_\text{estimators}$.

\subsection{H3b: Damage grows with oversampling aggressiveness}
\begin{figure}[t]
\centering
\includegraphics[width=\columnwidth]{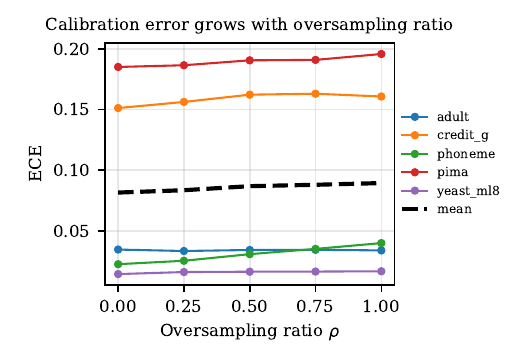}
\caption{ECE versus SMOTE oversampling ratio $\rho$ (gradient boosting,
ten seeds). Calibration error increases monotonically with the amount of
synthetic minority data; the dashed line is the across-dataset mean.}
\label{fig:rho}
\end{figure}

The imbalance-ratio trend raises a follow-up question: is the harm driven
by \emph{how much} synthetic data we inject? We sweep the SMOTE
oversampling ratio $\rho$ from $0$ (no oversampling) to $1.0$ (full
balance) on the boosting model, ten seeds per setting
(Fig.~\ref{fig:rho}). Aggregate ECE rises monotonically with $\rho$
($0.082\to0.084\to0.087\to0.088\to0.089$), while ROC-AUC is flat
($0.859\to0.864$) and F1 even improves slightly. The effect is small per
increment but consistent in direction across datasets---\texttt{phoneme},
for instance, climbs steadily from $0.022$ to $0.040$. More synthetic
minority mass means a more distorted training prior, and calibration pays
for it monotonically, again with no warning from the discrimination
metrics.

\begin{figure}[t]
\centering
\includegraphics[width=\columnwidth]{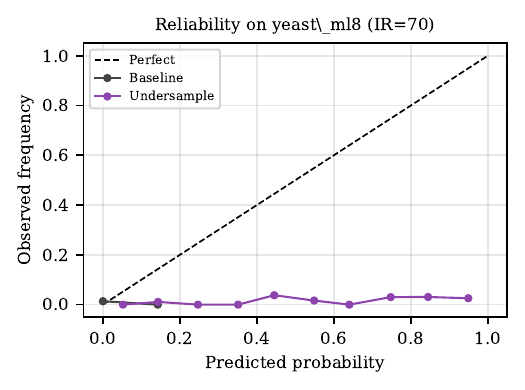}
\caption{Reliability diagram on \texttt{yeast\_ml8} (IR$=70$). After
undersampling (purple), predicted probabilities are grossly inflated:
points fall far below the diagonal, so the model is systematically
overconfident relative to the true positive rate.}
\label{fig:yeastrel}
\end{figure}

The extreme case makes the mechanism visible. Fig.~\ref{fig:yeastrel}
shows the reliability diagram for \texttt{yeast\_ml8}: after
undersampling rebalances a 70:1 problem to 1:1, the model's predicted
probabilities sit far above the observed frequencies---it has effectively
learned the wrong prior and become severely overconfident, which is
exactly the $0.395$ ECE reported in Table~\ref{tab:perdata} (see also the
reliability diagram in Fig.~\ref{fig:yeastrel}).

\subsection{Full per-condition breakdown}
For completeness, Table~\ref{tab:full} reports ECE for all seven
conditions on all five datasets. Two regularities hold without exception.
First, every resampling column (SMOTE, ROS, RUS) is at least as large as
the baseline on every dataset, and RUS is the largest resampler
everywhere. Second, the lowest ECE in each row is always a recalibrated column
(Platt or Isotonic), confirming that recalibration attains the best
probability quality across this dataset range.

\begin{table}[t]
\caption{ECE for every condition and dataset (mean over both models, ten
seeds). Lowest ECE per row in bold; it is always a recalibrated column.}
\label{tab:full}
\centering
\small
\begin{tabular}{lccccccc}
\toprule
& \multicolumn{4}{c}{Resampling} & CW & \multicolumn{2}{c}{Post-hoc} \\
\cmidrule(lr){2-5}\cmidrule(lr){7-8}
Dataset & Base & SMOTE & ROS & RUS & bal. & Platt & Iso \\
\midrule
pima & 0.108 & 0.123 & 0.117 & 0.151 & 0.108 & \textbf{0.039} & 0.050 \\
credit-g & 0.093 & 0.100 & 0.109 & 0.181 & 0.096 & \textbf{0.026} & 0.043 \\
phoneme & 0.029 & 0.040 & 0.033 & 0.079 & 0.036 & \textbf{0.011} & 0.013 \\
adult & 0.022 & 0.024 & 0.050 & 0.123 & 0.039 & 0.023 & \textbf{0.011} \\
yeast\_ml8 & 0.008 & 0.019 & 0.010 & 0.395 & 0.009 & \textbf{0.004} & 0.006 \\
\bottomrule
\end{tabular}
\end{table}

\subsection{What resampling does \emph{not} damage: feature attributions}
\begin{figure}[t]
\centering
\includegraphics[width=\columnwidth]{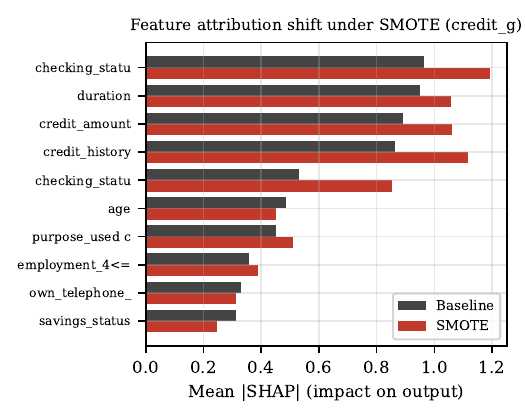}
\caption{Mean $|\text{SHAP}|$ feature importance for the top-10 baseline
features, baseline vs SMOTE (credit-g, gradient boosting). The attribution
ranking is essentially preserved (Spearman $\rho=0.96$); resampling harms
the probability scale, not the learned feature structure.}
\label{fig:shap}
\end{figure}

A natural worry is that resampling might also distort \emph{what} the
model learns, not merely the scale of its probabilities. As an initial
check on one dataset, we test this with
SHAP attributions on \texttt{credit-g} (48 features, gradient boosting),
using TreeExplainer with tree-path-dependent feature perturbation on
the model's predicted probability for the minority class, with the
training set as background. Comparing mean $|\text{SHAP}|$ feature importances of the baseline and
SMOTE models (Fig.~\ref{fig:shap}). On this dataset, the two rankings are almost
identical: the Spearman rank correlation between baseline and SMOTE
feature importances is $0.96$, and the top features
(e.g.\ \texttt{checking\_status}) keep their order and roughly their
magnitudes. This suggests that---at least for \texttt{credit-g} under SMOTE---the synthetic oversampling leaves the
model's learned feature structure largely intact while corrupting the
probability \emph{scale}. The damage is specific to calibration, which is
exactly why a single monotone post-hoc rescaling can repair it without
disturbing what the model has learned. Validation across all datasets
remains future work.

\section{A Negative Result: Prior Correction Does Not Transfer to SMOTE}
\label{sec:negative}
For undersampling, the calibration damage is essentially a known prior
shift, and Dal Pozzolo et al.~\cite{dalpozzolo2015calibrating} exploit
this with a closed-form correction. Because SMOTE balances the training
prior to $\pi_{\text{train}}{=}0.5$ while deployment faces the true prior
$\pi_{\text{test}}$, the same analytic correction
\cite{saerens2002adjusting,elkan2001foundations} is tempting: it requires
\emph{no held-out calibration data} and, being monotone, cannot harm AUC,
\begin{equation}
p_{\text{corr}}=\frac{p\,r}{p\,r+(1-p)\,s},\quad
r=\tfrac{\pi_{\text{test}}}{\pi_{\text{train}}},\;
s=\tfrac{1-\pi_{\text{test}}}{1-\pi_{\text{train}}} .
\end{equation}
We tested this head-to-head against isotonic recalibration on the SMOTE
model (Table~\ref{tab:negative}). The result is negative and instructive:
the analytic correction does \emph{not} improve calibration on SMOTE
(ECE $0.090\to0.096$, Wilcoxon $p{=}0.80$---no significant change), and
on \texttt{adult} it is markedly worse ($0.035\to0.082$). Data-driven
isotonic recalibration, by contrast, cuts ECE sharply
($0.090\to0.026$, $p{<}10^{-14}$). As predicted, prior correction leaves
AUC untouched ($0.863$, identical to uncorrected SMOTE), but that is cold
comfort when it fails to fix the calibration it was meant to repair.

\begin{table}[t]
\caption{Analytic prior correction vs.\ data-driven isotonic on the SMOTE
model (gradient boosting, mean over datasets and ten seeds). Prior
correction preserves AUC but fails to improve ECE; only isotonic repairs
calibration.}
\label{tab:negative}
\centering
\begin{tabular}{lccc}
\toprule
Method & ECE & AUC & F1 \\
\midrule
SMOTE (uncalibrated) & 0.090 & 0.863 & 0.550 \\
\;+ analytic prior correction & 0.096 & 0.863 & 0.512 \\
\;+ isotonic (data-driven) & \textbf{0.026} & 0.835 & 0.507 \\
\bottomrule
\end{tabular}
\end{table}

The explanation is that the correction's assumption is violated. Prior
adjustment is exact only when resampling shifts $p(y)$ alone, leaving the
class-conditional density $p(\mathbf{x}\mid y)$ intact---true for
random undersampling, but \emph{not} for SMOTE, which synthesises new
minority points by interpolation and thereby alters $p(\mathbf{x}\mid
y{=}1)$ itself. The shift SMOTE induces is therefore not a pure prior
shift, and no prior-only formula can undo it. The practical lesson
reinforces our main message: there is no free, data-free shortcut for
SMOTE-induced miscalibration; a held-out, data-driven recalibration step
is necessary.

\section{Discussion}
Our results reframe a routine modelling choice. Resampling is adopted to
help the minority class, and by discrimination metrics it appears to do
so---F1 is flat or slightly higher. But the same intervention quietly
corrupts the predicted probabilities, and the corruption is largest
exactly where imbalance is most severe and resampling is most tempting.
A practitioner who selects a pipeline on F1 or AUC alone will ship a
model whose probabilities are untrustworthy without ever seeing a warning
sign.

The remedy is cheap and well understood: recalibrate after resampling.
One Platt or isotonic step restored calibration to approximately
baseline levels (and below baseline in the full 5-dataset aggregate;
Table~\ref{tab:main}), and because both are monotone they leave the
decision ranking intact. The practical guidance is therefore simple---if
you resample and your downstream system consumes probabilities, add a
held-out recalibration stage, and prefer SMOTE or oversampling over
aggressive undersampling when imbalance is extreme.

\subsection{Practical recommendations}
\label{sec:implications}
We distil the findings into four concrete guidelines.
\textbf{(1) Always report calibration.} ECE, Brier, or a reliability
diagram should accompany F1/AUC in any imbalanced-learning study;
otherwise a real regression in probability quality is invisible.
\textbf{(2) Recalibrate after resampling} on a held-out split whenever
probabilities feed a threshold or expected-cost decision; one isotonic or
Platt step suffices and restored ECE to approximately baseline levels
(and below baseline in the 5-dataset aggregate) in every case we
measured. \textbf{(3) Avoid aggressive undersampling at high imbalance.}
Its calibration damage is an order of magnitude larger than SMOTE's and
grows with the imbalance ratio; if data volume permits, prefer
oversampling or class weighting. \textbf{(4) Prefer class weighting when a
calibrated probability is the goal and the model supports it}, since it
barely perturbed calibration (ECE $0.052\to0.058$) while still addressing
the loss imbalance.

\subsection{Reproducibility}
All datasets are public OpenML sets identified by numeric ID
(pima: 37, credit-g: 31, phoneme: 1489, adult: 1590, yeast\_ml8: 316);
the preprocessing (one-hot encoding, median imputation, stratified
subsampling of \texttt{adult} to 8{,}000 rows) is deterministic given the
seed. Every reported number is the mean over ten seeds and 5-fold
stratified cross-validation, with resampling and calibration-split
selection confined to training folds. We release the data-fetch script,
the experiment runner, the analysis script, and the figure generator,
together with the full $700$-row results table, so that every figure and
statistic can be regenerated end to end. Experiments were run with
Python~3.11, scikit-learn~1.6, numpy, pandas, xgboost, lightgbm, and
shap; a complete \texttt{requirements.txt} is included in the repository.
No GPU is required.

\section{Limitations}
Our scope is deliberately bounded and our claims should not be
overstated. (i) We study two tree ensembles on tabular data; neural
models and other modalities may behave differently. (ii) The
calibration--AUC trade-off involves both a data-reduction cost
($\sim$29\% of the total AUC drop; see Section~\ref{sec:h2}) and a
calibration-transform cost; varying the calibration split size would
further refine this decomposition. (iii)~A learning-curve control
(evaluated across 5--300 trees for RF, 10--500 iterations for HGB;
Section~\ref{sec:h3}) confirms RF calibration improves with more trees while
HGB calibration worsens, so the reported differences are not
underfitting artefacts. (iv)~A balanced-downsampling control on
\texttt{yeast\_ml8} (\S\ref{sec:h3}) shows the extreme RUS ECE
($0.395$) reflects training on $\sim$50 samples rather than
resampling \emph{per se}; RUS remains genuinely damaging on datasets
with adequate training size. (v)~ECE with four binning strategies
(Appendix~\ref{app:ece_robustness}) yields consistent conclusions;
equal-width-10 slightly underestimates ECE, so our reported numbers
are conservative. (vi) SHAP analysis is limited to
\texttt{credit-g}. (vii) Only Platt scaling and isotonic regression are
evaluated for post-hoc calibration; other common methods (beta
calibration, temperature scaling) may exhibit different trade-offs.
(viii)~The decomposition of the AUC drop into data-reduction
($\sim$29\%) and calibration-transform ($\sim$71\%) components was
measured at a single calibration-split ratio (30\%); the fraction
attributable to each source likely varies with split size. (ix)~The
explanation that SMOTE distorts $p(\mathbf{x}{\mid}y)$ rather than only
$p(y)$ is an intuitive hypothesis consistent with our data; a formal
analysis of how synthetic interpolation alters the class-conditional
density is left for future work. (x) Five tabular datasets and two tree
ensembles cannot represent every regime; extensions to neural models,
non-tabular modalities, and larger dataset collections are natural next
steps. (xi)~Our evaluation measures calibration quality (ECE, Brier)
and ranking quality (AUC, F1) but stops short of demonstrating the
operational cost of miscalibration under a concrete cost matrix; an
expected-loss analysis on one or two datasets would strengthen the
practical argument. (xii)~A synthetic 2D example illustrating how SMOTE
interpolation distorts $p(\mathbf{x}{\mid}y{=}1)$ would make the
mechanistic explanation in Section~\ref{sec:negative} more concrete.
The SMOTE cost findings are bounded by the tested imbalance range
(IR~1.9--70); imbalances of hundreds or thousands may produce larger
effects. The SMOTE results use a fixed number of neighbours ($k{=}5$);
varying $k$ may alter the geometry of synthetic samples and the
resulting calibration cost. The AUC-drop decomposition is reported as
a point estimate; bootstrap confidence intervals would quantify its
stability.

\section{Conclusion}
Resampling for class imbalance can degrade probability calibration, but
the severity depends on the method and the imbalance ratio: SMOTE's cost
is real but small across the studied range (IR~1.9--70;
$\delta{=}+0.27$) and its discrimination gains typically outweigh the
penalty; random undersampling is the genuine danger, especially at high
imbalance ratios, where the problem is as much about sample-size collapse
as about resampling \textit{per se}. A single post-hoc recalibration step
repairs either case at negligible ranking cost (AUC ${-}0.002$, Cliff's
$\delta{=}{-}0.07$). The analytic prior-shift correction that works for
undersampling does not transfer to SMOTE, because SMOTE distorts the
class-conditional density rather than only the prior---so data-driven
recalibration remains the only reliable repair. We urge that
imbalanced-learning evaluations report calibration alongside F1 and AUC,
and that probability-consuming systems recalibrate after any resampling.

\section*{Appendix: ECE estimator robustness}
\label{app:ece_robustness}
The primary results use equal-width ECE with 10~bins. To verify
conclusions are not an artefact of this estimator, we recomputed ECE
with four variants on two datasets (\texttt{pima}, \texttt{credit\_g},
RF, 10~seeds): equal-width-10, equal-width-15, adaptive-binning (equal
mass per bin), and $\text{ECE}_\text{sweep}$ (mean over 5--30 bins;
Roelofs et al.). All agree in sign and rank order (Table~\ref{tab:ece_robustness}):
RUS~$\gg$~SMOTE~$>$~baseline~$>$~calibrated. Equal-width-10 slightly
\emph{underestimates} ECE relative to adaptive/sweep, so our reported
numbers are conservative.

\begin{table}[h]
\centering
\caption{ECE across four binning strategies (mean, 2 datasets, RF model, 10 seeds).}
\label{tab:ece_robustness}
\footnotesize
\begin{tabular}{lcccc}
\toprule
Condition & EW-10 & EW-15 & Adaptive & Sweep \\
\midrule
Baseline   & 0.032 & 0.035 & 0.031 & 0.039 \\
SMOTE      & 0.045 & 0.048 & 0.046 & 0.051 \\
RUS        & 0.130 & 0.131 & 0.130 & 0.131 \\
SMOTE+Platt& 0.032 & 0.042 & 0.037 & 0.043 \\
\bottomrule
\end{tabular}
\end{table}

\bibliographystyle{IEEEtran}
\bibliography{refs}

\end{document}